\documentclass[sigconf]{acmart}

\usepackage{graphicx}
\usepackage[T1]{fontenc}
\usepackage{color}
\usepackage{textcomp}
\usepackage{tabulary}
\usepackage{url}
\usepackage{booktabs}
\usepackage{amsfonts}
\usepackage{nicefrac}
\usepackage{microtype}
\usepackage{booktabs}
\usepackage[utf8]{inputenc}
\usepackage{multirow}
\usepackage[export]{adjustbox}
\renewcommand{\paragraph}[1]{\vspace{-.2cm} \hfill \break \textbf{#1.}}
\usepackage{amsmath}
\usepackage{mathtools}
\usepackage{dsfont}
\usepackage{enumitem}
\usepackage{color}
\usepackage[caption=false]{subfig}
\usepackage{capt-of}
\usepackage{hyperref}

\newcommand{\xhdr}[1]{\vspace{0.2mm}\noindent{{\bf #1.}}}

\begin{document}

\copyrightyear{2018}
\acmYear{2018}
\setcopyright{iw3c2w3}
\acmConference[WWW 2018]{The 2018 Web Conference}{April 23--27, 2018}{Lyon,
France}
\acmBooktitle{WWW 2018: The 2018 Web Conference, April 23--27, 2018, Lyon, France}
\acmPrice{}
\acmDOI{10.1145/3178876.3186138}
\acmISBN{978-1-4503-5639-8/18/04.}

\fancyhead{}

\title{Human Perceptions of Fairness in Algorithmic Decision Making:\\ A Case Study of Criminal Risk Prediction}

\author{Nina Grgi\'{c}-Hla\v{c}a}
\affiliation{
 \institution{MPI-SWS, Saarland University}
}
 \email{nghlaca@mpi-sws.org}

\author{Elissa M. Redmiles}
\authornote{Elissa Redmiles acknowledges support from the National Science Foundation Graduate Research Fellowship Program under Grant No. DGE 1322106.}
\affiliation{
 \institution{University of Maryland}
}
 \email{eredmiles@cs.umd.edu}

\author{Krishna P. Gummadi}
\affiliation{
 \institution{MPI-SWS, Saarland University}
}
 \email{gummadi@mpi-sws.org}

\author{Adrian Weller}
\authornote{Adrian Weller acknowledges support from the David MacKay Newton research fellowship at Darwin College, The Alan Turing Institute Institute under EPSRC grant EP/N510129/1 \& TU/B/000074, and the Leverhulme Trust via the CFI.}
\affiliation{
 \institution{Cambridge University, Alan Turing Institute}
}
 \email{adrian.weller@eng.cam.ac.uk}

\renewcommand{\shortauthors}{Grgi\'{c}-Hla\v{c}a et al.}


\begin{abstract}

As algorithms are increasingly used to make important decisions that
affect human lives, ranging from social benefit assignment to
predicting risk of criminal recidivism,
concerns have been raised about the fairness of algorithmic decision
making.
Most prior works on algorithmic fairness {\it normatively} prescribe
how fair decisions ought to be made. In contrast, here, we {\it
  descriptively} survey users for how they perceive and reason about
fairness in algorithmic decision making.

A key contribution of this work is the framework we propose to
understand \textit{why people perceive certain features as fair or
  unfair to be used in algorithms}.  Our framework identifies eight
properties of features, such as \textit{relevance},
\textit{volitionality} and \textit{reliability}, as latent
considerations that inform
people's moral judgments about the fairness of feature use in
decision-making algorithms.  We validate our framework through a
series of scenario-based surveys with 576 people.  We find that, based
on a person's assessment of the eight latent properties of a feature
in our exemplar scenario, we can accurately (> 85\%) predict if the
person will judge the use of the feature as fair.

Our findings have important implications. At a high-level, we show
that people's unfairness concerns are multi-dimensional and argue that
future studies need to address unfairness concerns beyond
discrimination. At a low-level, we find considerable disagreements in
people's fairness judgments. We identify root causes of
the disagreements, and note possible pathways to resolve them.

\end{abstract}

\maketitle

\section {Introduction} \label{sec:intro}

Algorithms trained over data about past decisions are increasingly
used to assist or replace human decision making in life-affecting
scenarios, such as determining if an unemployed person should be
eligible for a certain level of social welfare
benefits~\cite{polish_unemployment} or deciding if a person should be
let out on bail pending trial~\cite{propublica_story}. Given their
potential impact on human lives, concerns have been raised about the
\textit{fairness} of the decisions made by
algorithms~\cite{agan2016ban,propublica_story,barocas_2016,flores_propublica_reply}.

Concerns about algorithmic unfairness have led to much recent work on
detecting and mitigating {\it discrimination} in decision-making
scenarios. This work includes finding ways to operationalize notions
of direct and indirect discrimination and provide
mechanisms~\cite{Dwork2012,feldman_kdd15,kamiran_sampling,salvatore_knn,pedreschi_discrimination,zafar_fairness,zafar_dmt,icml2013_zemel13}
for non-discriminatory learning, as well as examining the feasibility
of making non-discriminatory
decisions~\cite{dimpact_fpr,goel_cost_fairness,
  friedler_impossibility, kleinberg_itcs17}.

Existing studies of algorithmic fairness are largely {\it normative
  (prescriptive)} in nature, i.e., they begin by defining how fair
decisions should (or ought to) be made, assuming that there is
societal consensus around what constitutes fair decision
making~\cite{sep-morality-definition}.
In this paper, we pursue a complementary {\it descriptive
  (comparative)} approach towards fair decision making. Inspired by
works in descriptive ethics~\cite{sep-morality-definition}, we conduct
empirical studies in one specific context, to learn what people
perceive as fair decision making, with the goal of uncovering the
moral reasoning behind their perceptions. Later, we discuss how our
findings can be leveraged to design fair decision-making algorithms.

As perceptions of fairness are {\it multi-dimensional} and {\it
  context-dependent}, characterizing them presents a difficult
challenge. In this work, we propose to understand how people make
judgments about the fairness of using individual features in decision
making. More concretely, we seek to measure and analyze how people
would answer the following question: {\bf Is it fair to use a feature
  ($\mathcal{F}$) in a given decision making scenario
  ($\mathcal{S}$)?}

We center our investigation of fairness perceptions around the above
question for multiple reasons: First, people's judgments about
fairness of using features can be leveraged to learn fair algorithmic
decision making, as shown in our recent work \cite{grgic2016case,
  grgic2018beyond}.  Second, while the question, ``is this feature
fair to use,'' is intuitive and simple to comprehend, people's answers
(as our study shows) can be analyzed to reveal the extent to which
different types of fairness considerations, such as whether or not a
given feature is {\it volitional} or {\it causes the outcome}, factor
into their judgments.

\vspace{3mm}
\xhdr{Our Contributions}
We collected and analyzed fairness judgments from a survey of 576
people. We asked
survey
participants to assess the fairness of
using different features that are inputs to
COMPAS~\cite{propublica_story}, a commercial criminal risk estimation
tool that is in current use to help make judicial decisions in the US about bail decisions.\footnote{COMPAS may also be used for criminal sentencing and parole decisions, but we do not explore those uses here since they involve other relevant factors such as considering the role of long term incarceration in society.}

To model the factors that drive participants' fairness judgments, we
propose a set of eight {\it latent properties} of features that we
hypothesize capture most of the considerations that influence
people's fairness judgments. Our framework of eight properties
includes unfairness concerns beyond discrimination, such as whether a
feature is privacy sensitive and whether it is volitional (see Section \ref{sec:framework}).

When asked to assess the fairness of using the different input
features to COMPAS for making bail decisions, the majority of our respondents judged that half
 of the features are unfair to use. Interestingly, the latent
properties the respondents' considered in reaching their judgments
were mostly unrelated to discrimination, highlighting the need to
consider additional unfairness concerns.

Unfortunately, we also find that there is a lack of clear consensus in
respondents' judgments about the fairness of using a number of
features. Our analysis attempts to explain the lack of consensus by
modeling people's fairness judgments as a {\it two-part} decision
process: one related to how people assess the latent properties of a
feature, and a second related to how they morally reason about the
fairness of using a feature given that is has certain latent properties.

We find that the lack of consensus in our respondents' fairness
judgments can be largely attributed to disagreements in how they
estimated the latent properties --
particularly
those related to causal reasoning, such as whether a feature causes
the outcome or is caused by sensitive group membership. However, we
find that respondents use similar moral reasoning in reaching
fairness judgments if given a set of latent properties. Specifically, we were able to learn a single, simple
classifier which performs accurately in predicting respondents' fairness judgments from their latent property assessments.

\vspace{3mm}
\xhdr{Implications} Our findings are striking because they suggest
that the differences in people's fairness judgments may originate not
from the differences in their inherently {\it more subjective} moral
reasoning about how to weigh different latent properties
when judging fairness of features, but rather from differences in their
seemingly {\it more objective} assessments of latent properties of
features.  As such, our findings
point towards a future for fair algorithmic decision making where the latent
properties of a feature (e.g., whether it causes the outcome) might be
objectively determined from extensive data,
while there is hope that the moral reasoning that people use to map the latent
properties to fairness might be consistently determined using input from people
(e.g., collected via surveys).

\section{Judging Feature Usage Fairness}
\label{sec:framework}
Many works in psychological decision theory propose that people use
heuristics to assess situations and reach
decisions~\cite{einhorn1981behavioral, albar2009heuristics}. These
heuristics may vary according to the situation and to people's level of knowledge about
the situational elements, helping to parse information into more
manageable and meaningful pieces~\cite{gigerenzer2011heuristic}.

We hypothesize that when determining whether a feature is fair to be
used in a decision making scenario, people rely on their implicit or
explicit assessments of certain
underlying properties of the feature as a heuristic. So our framework
for how people judge feature usage fairness consists of two-parts. In
the first part, we conjecture eight latent properties of a given
feature as potential determinants for how people judge the fairness of using the feature. In the
second part, we hypothesize that these latent properties are weighted in
different ways by different individuals when reaching their fairness
judgments about the feature.
We draw these latent properties from the existing literature in
social-economic-political-moral sciences, philosophy, and the law, as
detailed below.

\xhdr{I. Reliability} Inspired by legal requirements that any
admissible evidence be {\it reliably assessed}~\cite{Kumho1999, Daubert1993},
fairness judgments might be
influenced by the potential for reliably assessing the feature. For
instance, an input feature to COMPAS recidivism risk prediction is
the defendant's belief about criminality, assessed via answers to
questions of the form ``Do you think that a hungry person has a right
to steal?'' People who do not perceive beliefs about criminality as
reliably assessable from such questions might rate the feature as
unfair to use.

\xhdr{II. Relevance} Inspired by legal requirements that any admissible
evidence be {\it relevant} to the case~\cite{fre401,fre402},
fairness judgments might be
influenced by a feature's relevance to the decision making scenario. For
instance, an input feature to COMPAS recidivism risk prediction is the
defendant's education and behavior in school, assessed via answers to
questions of the form ``What were your usual grades in high school?''
People who perceive performance in school as irrelevant to recidivism
risk estimation might rate the feature as unfair to use.

\xhdr{III. Privacy} Inspired by legal requirements that evidence
obtained via illegal privacy intrusions (such as searches without
warrant or unauthorized wire tapping) is inadmissible ~\cite{fre402,Katz1967,Weeks1914},
fairness judgments might be influenced by whether the feature is relying on
privacy-sensitive information. For instance, an input feature to
COMPAS recidivism risk prediction is the defendant's history of substance
abuse, assessed via answers to questions of the form ``Did you use
heroin, cocaine, crack, or meth as a juvenile?''  People who perceive
juvenile drug use as privacy-sensitive information might rate the
feature as unfair to use.

\xhdr{IV. Volitionality} Inspired by philosophical arguments on luck
egalitarianism~\cite{anderson1999point,arneson2000luck,naudts2017fair} that people should be held responsible for the
voluntary choices they make (option luck), but not penalized for their
unchosen circumstances (brute luck),
fairness judgments might be influenced by
an individual's assessment of the extent to which a feature is {\it
  volitional}, i.e., the result of exercising one's own will. For
instance, an input feature to COMPAS recidivism risk prediction is the
criminal history of the defendant's family, assessed via answers to
questions of the form ``Was your father or mother ever arrested?''
People who perceive family criminal history as non-volitional might
rate the feature as unfair to use.

\xhdr{V. Causes Outcome} Inspired by arguments for applying causal
reasoning in fairness~\cite{bonchi2017exposing,kilbertus2017avoiding,kusner2017counterfactual,whenworlds,nabi2017fair,qureshi2016causal,zhang2017anti},
fairness judgments might be influenced by whether a
feature is likely to cause (i.e., increase or mitigate) the chances of the
person engaging in risky behavior. For instance, an input feature to
COMPAS recidivism risk prediction is the defendant's current charge,
assessed via answers to questions of the form ``Are you currently
charged with a misdemeanor, non-violent felony or violent felony?''
People who perceive the defendant's current charge as causing him to
recidivate might consider the feature fair to use.

\xhdr{VI. Causes Vicious Cycle} Inspired by arguments for avoiding
vicious cycles of crime and poverty~\cite{gallie2003unemployment,mosley2005risk},
fairness judgments might be influenced by
whether a feature is likely to trap people in a
vicious cycle of increasingly risky behaviors. For instance, an input
feature to COMPAS recidivism risk prediction is the criminal history
of the defendant's friends, assessed via answers to questions of the
form ``How many of your friends have ever been arrested?''  People who
perceive friends' criminal history
may create a vicious
cycle (where people with friends with criminal records are sentenced
to longer prison terms, thereby, increasing the number friends with
criminal records) might rate the feature as unfair to use.

\xhdr{VII. Causes Disparity in Outcomes} Inspired by the doctrine of
disparate impact in anti-discrimination laws that require statistical
parity in outcomes for people belonging to different sensitive social
groups like race or gender~\cite{civil_rights_act,barocas_2016},
fairness judgments might be influenced by
whether a feature would result in protected group members receiving
disadvantageous treatment. For instance, an input feature to COMPAS
recidivism risk prediction is the safety of the neighborhood the
defendant is living in, assessed via answers to questions of the form
``Is there much crime in your neighborhood?''  People who perceive
neighborhood safety
may increase disparity in
outcomes might rate the feature as unfair to use.

\xhdr{VIII. Caused by Sensitive Group Membership} Inspired by the notions
of indirect discrimination in political and economic sciences, where
members of a social group are implicitly discriminated against using
features that are correlated with or caused by their group
membership~\cite{civil_rights_act,barocas_2016},
fairness judgments might be influenced by
the extent to which
a feature is caused by their group membership. For instance, an
input feature to COMPAS recidivism risk prediction is the criminal
history of the defendant's friends, assessed via answers to questions
of the form ``How many of your friends have ever been arrested?''
People who perceive friends' criminal history as potentially caused by
people's membership of some social groups might rate the feature as
unfair to use.

\xhdr{Observation 1: Sufficiency and Necessity of our Latent
  Properties} We do not claim that our list of latent properties
presented above is {\it exhaustive or complete}. That is, there may
exist other properties that might influence users' fairness
judgments. However, as we show in
Section~\ref{subsec:pil1}, the eight properties are by and
large {\it sufficient and necessary} to explain fairness judgments of
users in our survey. Specifically, less than 3\% of our surveyed users
reported using a property outside of our list in arriving at their
judgments; further, for each of the eight properties, at least 15\% of our surveyed users reported relying on it as a consideration in their fairness
judgments.
Moreover, when we attempted analytically to predict
users' fairness judgments based only on their assessments of the latent
properties in Section~\ref{subsec:fairness_prediction}, we found that the eight properties are not only sufficient
to make the predictions with high accuracy, but that six of the eight
are also statistically significant (i.e., necessary) for predicting
fairness judgments.

\xhdr{Observation 2: Unfairness beyond Discrimination} Our list of
latent properties captures a diverse set of unfairness concerns with
algorithmic decision making that go beyond {\it discrimination}, the
traditional basis for most of the existing literature on algorithmic
fairness. In fact, the two properties that were not deemed
statistically significant in our prediction analysis discussed above,
\textit{Causing Disparity in Outcomes} and \textit{Caused by Sensitive
  Group Membership}, are related to the potential for a feature to
cause discrimination. Thus, our proposed framework captures many
facets of unfairness
that have previously received little attention in the fair learning community, yet may
significantly influence users' fairness perceptions of algorithmic
decision making.

\begin{table*}[ht]
\centering
\begin{tabulary}{\linewidth}{LL|L}
  \hline
  &\textbf{Predictive Feature} & \textbf{Example Question}\\
\hline
  \textbf{1.} & Current Charges & Are you currently charged with a misdemeanor, non-violent felony or violent felony?\\
  \textbf{2.} & Criminal History: self & How many times have you violated your parole? \\
  \textbf{3.} & Substance Abuse & Did you use heroin, cocaine, crack or meth as a juvenile? \\
  \textbf{4.} & Stability of Employment \& Living Situation & How often do you have trouble paying bills?\\
  \textbf{5.} & Personality & Do you have the ability to ``sweet talk'' people into getting what you want? \\
  \textbf{6.} & Criminal Attitudes & Do you think that a hungry person has a right to steal? \\
  \textbf{7.} & Neighborhood Safety & Is there much crime in your neighborhood? \\
  \textbf{8.} & Criminal History: family and friends &  How many of your friends have ever been arrested?\\
  \textbf{9.} & Quality of Social Life \& Free Time & Do you often feel left out of things?\\
  \textbf{10.} & Education \& School Behavior & What were your usual grades in high school?\\

\hline
\end{tabulary}
\caption{\small The ten features assessed in our survey and the questions provided as examples in the scenario. The features and questions are drawn from the COMPAS questionnaire.}
\label{table:features}
\vspace{-4ex}
\end{table*}

\section{Methodology}\label{sec:methodology}
In order to gather people's judgments about algorithmic fairness and
the latent properties that we proposed, we conducted a series of
online surveys in September and October 2017. Our methodology was
approved by our institution's ethics review board.

\subsection{Survey Design}
We asked participants to respond to questions in the context of a
specific scenario that is already in use in the real world.

\subsubsection{\bf Scenario}
We consider the COMPAS (Correctional Offender Management Profiling for
Alternative Sanctions) system, which analyzes defendant's answers to a
large questionnaire with questions across a range of categories in
order to predict risk of criminal activity. COMPAS has been adopted
across various jurisdictions in the US
to assist with tasks
from the judicial domain, including decisions about bail, sentencing
lengths and parole \cite{propublica_story}.

Our survey begins with the following: ``Judges in Broward County,
Florida, have started using a computer program to help them decide
which defendants can be released on bail before trial. The computer
program they are using takes into account information about
\textbf{<feature>}.  For example, the computer program will take into
account the defendant's answer to the following question:
\textbf{<question>}.''  These items were asked for ten features
related to the COMPAS tool, outlined in
Table~\ref{table:features}.\footnote{For a full description of the
  features, as used in the survey, please see
  \url{https://fate-computing.mpi-sws.org/procedural_fairness/}} These
features were drawn from the categories in the COMPAS questionnaire,
and the example question for that feature was extracted from the
corresponding category in the
questionnaire~\cite{compas_questionnaire}.  We use this scenario in
two pilot surveys and the main survey, as discussed
below.

\subsubsection{\bf Pilot Survey 1: Fairness Judgments and Their Latent Reasons}\label{subsec:pil1}
In \textit{Pilot survey 1}, we sought to learn whether respondents
found the above scenario \textit{fair}, and \textit{why they felt it
  was fair} or unfair.

We asked people to assess whether the scenario was fair on a 7-point Likert scale from ``Strongly Disagree'' to ``Strongly Agree''.
Specifically, we asked: ``Please rate how much you agree with the following statement: It is fair to determine if a person can be released on bail using information about their \textbf{<feature>}.''
Then we asked them to select their reasons for why it was fair or unfair, providing the eight latent properties as answer options (as described below, for \textit{Pilot survey 2}) , while also providing an ``Other'' option with a text-entry box.
Participants who rated the scenario as unfair (from 1-3) were only asked why it was unfair, and participants who rated the scenario as fair (from 5-7) were only asked why it was fair. Participants who gave the scenario a neutral rating (4) were asked both questions.
Participants were asked this set of questions for each of the ten features in Table~\ref{table:features}, with the order in which the features were presented between respondents randomized.

\begin{figure}[t]
\centering

\includegraphics[width=0.9\columnwidth]{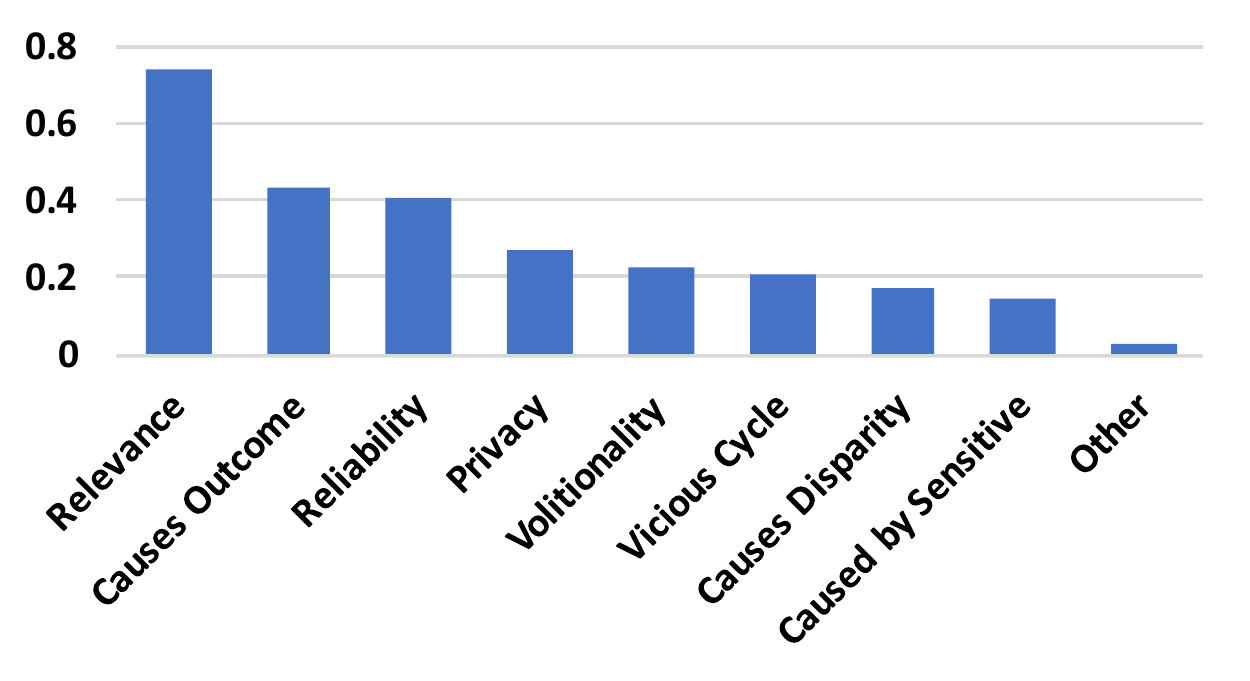}

\vspace{-4ex}

\caption{\small Properties used as justifications of fairness judgments in \textit{Pilot survey 1}. For each property, the plot shows the percentage of responses that used it as a justification of the fairness judgment. Note that multiple properties can be used as a justification for a single judgment.}
\label{fig:properties:used}
\vspace{-4ex}
\end{figure}
\xhdr{Takeaways} We used {\it Pilot survey 1} to assess whether the
properties we propose in Section~\ref{sec:framework} are
\textit{necessary} and \textit{sufficient} to capture people's
reasoning about fairness.  We find that each property was used by at
least 15\% of respondents to explain why they rated a particular
feature as fair or unfair to use
(Figure~\ref{fig:properties:used}). Interestingly, the properties that
were used as explanations of fairness judgments the most frequently
are not related to notions of discrimination: \textit{Relevance} is
the most used property, used in 74\% of the responses, followed by
\textit{Causes Outcome} and \textit{Reliability}, used in more than
40\% of the responses.

Further, less than 3\% of respondents selected the ``Other'' option
for why they judged a scenario as fair or unfair. Thematic analysis~\cite{braun2006using} of responses provided in the ``Other'' category reveals that the majority of these responses still map to one of our eight properties.
The frequent selection of each of our proposed properties, and the low proportion of ``Other'' responses, suggest that we are unlikely to be missing an important assessment criterion.

\subsubsection{\bf Pilot Survey 2: Latent Properties of Features}\label{subsec:pil2}
In \textit{Pilot survey 2} we sought to explore how people evaluate
the latent properties of features from our framework.  Here we asked
no fairness-related questions, in order to control for the effect of
asking about fairness on latent property ratings, as discussed in
Section~\ref{sec:meth:limitations}.

We presented the scenario and asked people to assess the value of the eight \textit{properties} of the features on the same 7-point Likert scale.
The properties were described as follows:
\begin{description}
\item [\textbf{I.} Reliability:] Information about <feature> can be assessed reliably.
\item [\textbf{II.} Relevance:] Information about <feature> is relevant for making this decision.
\item [\textbf{III.} Privacy:] Information about <feature> is private.
\item [\textbf{IV.} Volitionality:] A person can change <feature> by making a choice or decision.
\item [\textbf{V.} Causes Outcome:] <feature> can cause them to breach their bail.
\item [\textbf{VI.} Causes Vicious Cycle:] Making this decision using information about <feature> can cause a vicious cycle.
\item [\textbf{VII.} Causes Disparity in Outcomes:] Making this decision using information about <feature> can have negative effects on certain groups of people that are protected by law (e.g., based on race, gender, age, religion, national origin, disability status).
\item [\textbf{VIII.} Caused by Sensitive Group Membership:] <feature> can be caused by their belonging to a group protected by law (e.g., race, gender, age, religion, national origin, disability status).
\end{description}
The order in which the features and latent properties were presented was randomized between respondents.

\xhdr{Takeaways} As we ask no fairness-related questions in {\it Pilot
  survey 2}, we use it to examine latent property assessments
independent of fairness. The results of this survey can help us
understand the bias that is introduced by asking about both latent
properties and fairness. To quantify this bias, we compare the results
of this survey to those generated from our \textit{Main survey},
described below, which included both latent property rating and
fairness rating questions. The details of this comparison are
discussed in Section~\ref{sec:meth:limitations}.

\subsubsection{\bf Main Survey: Fairness Judgments and Latent Properties of Features} \label{subsec:mainsurv}
In the \textit{Main survey} we sought to evaluate whether people's
judgments about the latent properties of features proposed in our
framework were relevant to their judgments about the fairness of
features.

In the \textit{Main survey}, we asked people questions about (i) the
fairness of features, as in the first question in \textit{Pilot survey
  1}, as well as about (ii) their latent property assessments, as in
\textit{Pilot survey 2}.  As in the pilot surveys, in the \textit{Main
  survey}, this set of fairness and latent property assessment
questions was asked for \textit{each} of the ten features from
Table~\ref{table:features}, with the order of features and latent
properties being randomized between respondents.  Additionally, we
randomized whether the fairness question was presented before or after
the questions about the latent properties.

\subsubsection{\bf Questionnaire Validity}
\label{sec:meth:validity}
To ensure that survey participants can meaningfully interpret our
questions, we pre-tested all items in the questionnaires using
cognitive interviews. Cognitive interviews involve asking participants
to think aloud to the researcher as they take the survey. This
approach is a survey methodology best-practice for ensuring construct
validity and questionnaire
accuracy~\cite{coginterview:willis,Presser01032004}. We conducted
cognitive interviews with five demographically diverse participants,
recruited using the Prolific crowdsourcing platform, and iteratively
refined our questionnaires based on participant feedback until new
considerations stopped emerging.

Once we were satisfied with the validity of our questionnaire, we
collected our final sample for analysis. To mitigate the effects of
order bias~\cite{redmiles2017summary}, we randomized the order of
questions described in
Sections~\ref{subsec:pil1}-\ref{subsec:mainsurv}.  Finally, we
included an attention-check question to ensure that participants were
thoughtfully answering the questionnaire items~\cite{krosnick2010}.

\subsection{Survey Samples and their Demographics}
\label{sec:meth:sample}
The \textit{Main survey} samples consisted of 196
Amazon Mechanical Turk (AMT) \textit{master workers} from the US and
380 US respondents with census-representative demographics, collected
using the survey recruitment firm Survey Sampling International (SSI).

We sampled users from two different platforms in our main survey,
because we were concerned about both representativeness and quality of
user responses. AMT users are known to provide responses of equal or,
often, higher quality than survey panel
respondents~\cite{dupuis2013analysis,buhrmester2011amazon}. However,
AMT workers are not demographically representative of the U.S.
population
due to selection bias introduced by differences between those who sign
up for AMT and the general population~\cite{redmiles2017summary}. SSI
and other such sampling firms, on the other hand, use a myriad of
different recruitment mechanisms to reduce selection bias and ensure
that a demographically-representative sample is recruited.

\begin{table}[ht]
\centering
\small
\begin{tabulary}{\linewidth}{L|R|R|R}
\hline
{\bf Demographic Attribute} & {\bf AMT} &{\bf SSI} & {\bf Census}\\
\hline
Male & 55\% & 44\% & 49\%\\
Female & 43\% & 55\% & 51\%\\
\hline
African American & 9\% & 12\% & 13\% \\
Asian & 3\% & 4\% & 6\% \\
Caucasian & 76\% & 71\% & 61\%\\
Hispanic & 8\% & 11\% & 18\% \\
Other & 2\% & 2\% & 4\% \\
\hline
$<$B.S. & 47\% & 68\% & 70\% \\
B.S.+ & 51\% & 32\% & 30\%\\
\hline
Liberal & 57\% & 37\% & 33\%*\\
Conservative & 17\% & 24\% & 29\%*\\
Moderate & 21\% & 33\% & 34\%*\\
Other  & 5\% & 6\% & 4\%*\\
\hline
\end{tabulary}
\caption{\small Demographics of our AMT and SSI survey samples
  compared to the 2016 U.S. Census~\cite{census:acs}. Figures marked
  with a * were compared to Pew data~\cite{pew:politics} for political
  leaning.}
\label{table:demographics}
\vspace{-6ex}
\end{table}

Table~\ref{table:demographics} shows the demographics of our AMT and
SSI survey samples, compared with the 2016
U.S. Census~\cite{census:acs}.
We find that our AMT respondents consisted of considerably fewer
females (43\%), more Caucasians (76\%), more highly educated
individuals (51\% hold at least a college degree), and more liberals
(57\%) than the U.S. population. This educational skew in AMT
population is consistent with observations from previous research
studies~\cite{ipeirotis2010demographics, ross2010crowdworkers,
  paolacci2010running}. On the other hand, our SSI respondents were
census-representative to within 5\% along the demographics of gender
(55\% female), education (32\% with a BS or above), and political
leaning (37\% liberal). While more of our SSI respondents identify as
Caucasian (71\%) than in the U.S. population, a possible explanation could be that respondents may have selected only Caucasian rather than both Hispanic and Caucasian in response to our race and ethnicity question, since it was not multiple-option.

\begin{table*}[ht!]
\centering
\begin{tabulary}{\linewidth}{LL|c|ccccccccc|cc}
  \hline
   & & & \multicolumn{9}{c|}{\textbf{Fraction of People Rating Feature}} & \multicolumn{2}{c}{\textbf{Consensus}} \\
  & & \textbf{Mean} & \multicolumn{4}{c}{\textbf{Unfair}} & & \multicolumn{4}{c|}{\textbf{Fair}} & \multicolumn{2}{c}{\textbf{1 - NSE}} \\
  \multicolumn{2}{c|}{\textbf{Feature}} & \textbf{fairness} & \textbf{1} & \textbf{2} & \textbf{3} & \textbf{1-3} & \textbf{4} & \textbf{5-7} & \textbf{5} & \textbf{6} & \textbf{7} & \textbf{7 pt} & \textbf{3 pt} \\
\hline
 \textbf{1.} & Current Charges & 6.38 & 0.01 & 0.01 & 0.01 & 0.03 & 0.03 & \textbf{0.95} & 0.12 & 0.18 & 0.65 & 0.46 & 0.78\\
 \textbf{2.} & Criminal History: self & 6.37 & 0.02 & 0.01 & 0.01 & 0.03 & 0.03 & \textbf{0.94} & 0.08 & 0.22 & 0.64 & 0.45 & 0.75\\
 \textbf{3.} & Substance Abuse & 4.84 & 0.08 & 0.07 & 0.10 & 0.24 & 0.07 & \textbf{0.68} & 0.26 & 0.22 & 0.20 & 0.07 & 0.28\\
 \textbf{4.} & Stability of Employment & 4.49 & 0.13 & 0.05 & 0.11 & 0.29 & 0.09 & \textbf{0.62} & 0.26 & 0.24 & 0.12 & 0.06 & 0.20\\
 \textbf{5.} & Personality & 3.87 & 0.16 & 0.18 & 0.11 & \textbf{0.44} & 0.10 & \textbf{0.46} & 0.22 & 0.12 & 0.12 & 0.02 & 0.14\\
 \textbf{6.} & Criminal Attitudes & 3.63 & 0.22 & 0.12 & 0.16 & \textbf{0.51} & 0.09 & \textbf{0.40} & 0.20 & 0.11 & 0.09 & 0.03 & 0.15\\
 \textbf{7.} & Neighborhood Safety & 3.14 & 0.28 & 0.21 & 0.15 & \textbf{0.64} & 0.07 & 0.30 & 0.12 & 0.10 & 0.08 & 0.06 & 0.25\\
 \textbf{8.} & Criminal History: family and friends & 2.78 & 0.38 & 0.21 & 0.09 & \textbf{0.67} & 0.07 & 0.26 & 0.13 & 0.10 & 0.03 & 0.13 & 0.27\\
 \textbf{9.} & Quality of Social Life \& Free Time & 2.70 & 0.38 & 0.20 & 0.12 & \textbf{0.70} & 0.07 & 0.23 & 0.12 & 0.08 & 0.03 & 0.13 & 0.29\\
 \textbf{10.} & Education \& School Behavior & 2.70 & 0.34 & 0.22 & 0.14 & \textbf{0.71} & 0.08 & 0.21 & 0.13 & 0.06 & 0.03 & 0.12 & 0.29\\

\hline
\vspace{-4ex}
\end{tabulary}

\caption{\small People's judgments on the fairness of using features, and the consensus in their responses, for the AMT sample.
The reported values of consensus are calculated as 1 - Normalized Shannon Entropy (NSE) of the responses.
In the 7 point column, we report consensus across the whole range of responses. In the 3 point column, we report consensus across responses bucketed into three main fairness categories: unfair (1-3), neutral (4) and fair (5-7).}
\label{table:fairness:judgments}
\vspace{-4ex}
\end{table*}

\subsection{Analysis Methods}\label{subsec:method_analysis}
In our analysis we measure the consensus amongst people's ratings of fairness and latent property values using Shannon entropy \cite{shannon1948mathematical} calculated over the probability distributions over the answers.
Shannon entropy \cite{frey1984consensus, alston1992there} and measures derived from Shannon entropy \cite{tastle2007consensus, herrera2011consensual} are frequently used to quantify consensus.
We calculate the Shannon Entropy normalized between 0 and 1 ($NSE$), and report values of consensus calculated as $1 - NSE$, so that complete consensus corresponds to 1 and complete disagreement to 0.

We also examine the predictive power of our framework by building a binary classifier that predicts if a feature will be considered fair (completely, mostly, slightly, neutral) or unfair (completely, mostly, slightly), based on the values of its latent properties.
The training data consists of respondents' evaluations of latent properties, and binarized evaluations of fairness.
We train a logistic regression model with $L2$ regularization, implemented with the Python Scikit-learn package \cite{scikit-learn}.
To evaluate the model, we randomly split the data into 50\%/50\% train/test folds five times, and report the average accuracy and AUC.
Further, we randomly select one of the five runs and analyze its missclassiffications on the whole data; the other runs yielded qualitatively similar results.

\subsection{Discussion of Limitations}
\label{sec:meth:limitations}
As in all survey studies, self-report biases may affect the data. As described in Section~\ref{sec:meth:validity} we have tried to mitigate these self-report biases as much as possible through extensive pre-testing and the adoption of best practices for question randomization.

As noted above, we used \textit{Pilot survey 2} to measure the amount of bias introduced into our data by asking about fairness judgments in the same survey as we ask about the latent properties. We calculate the probability distribution over latent property ratings for (i) \textit{Pilot survey 2} (the control), where we do not ask about fairness, and (ii) the \textit{Main survey}, where we ask about fairness {\it and} latent properties.
We see that the KL-divergence \cite{kullback1951information} from (i) to (ii) is very low, achieving values below 0.1 for $90\%$ of the questions, and below 0.14 for the remaining $10\%$.
We interpret these results, in accordance with \cite{burnham2003model}, as showing that if (i) is the real distribution of people's assessments, (ii) is a good approximation of it.
Therefore, we conclude that assessments of latent property values are minimally affected by questions about fairness.

Self-report studies may also suffer from generalizability biases, as those who take these surveys may not be representative of the general population. In order to maximize both generalizability and data quality we recruit both a census-representative population and an AMT sample, as detailed in Section~\ref{sec:meth:sample}. Finally, it is possible that people may feel differently about fairness in different contexts. Future work should seek to validate whether a model that produces actual results based on self-report inputs such as ours aligns with ``real-world'' fairness perceptions when people are placed in more ecologically-valid situations with decision-making algorithms.

\begin{figure*}[ht]
\centering

\subfloat{\includegraphics[height=0.95\columnwidth,angle=-90]{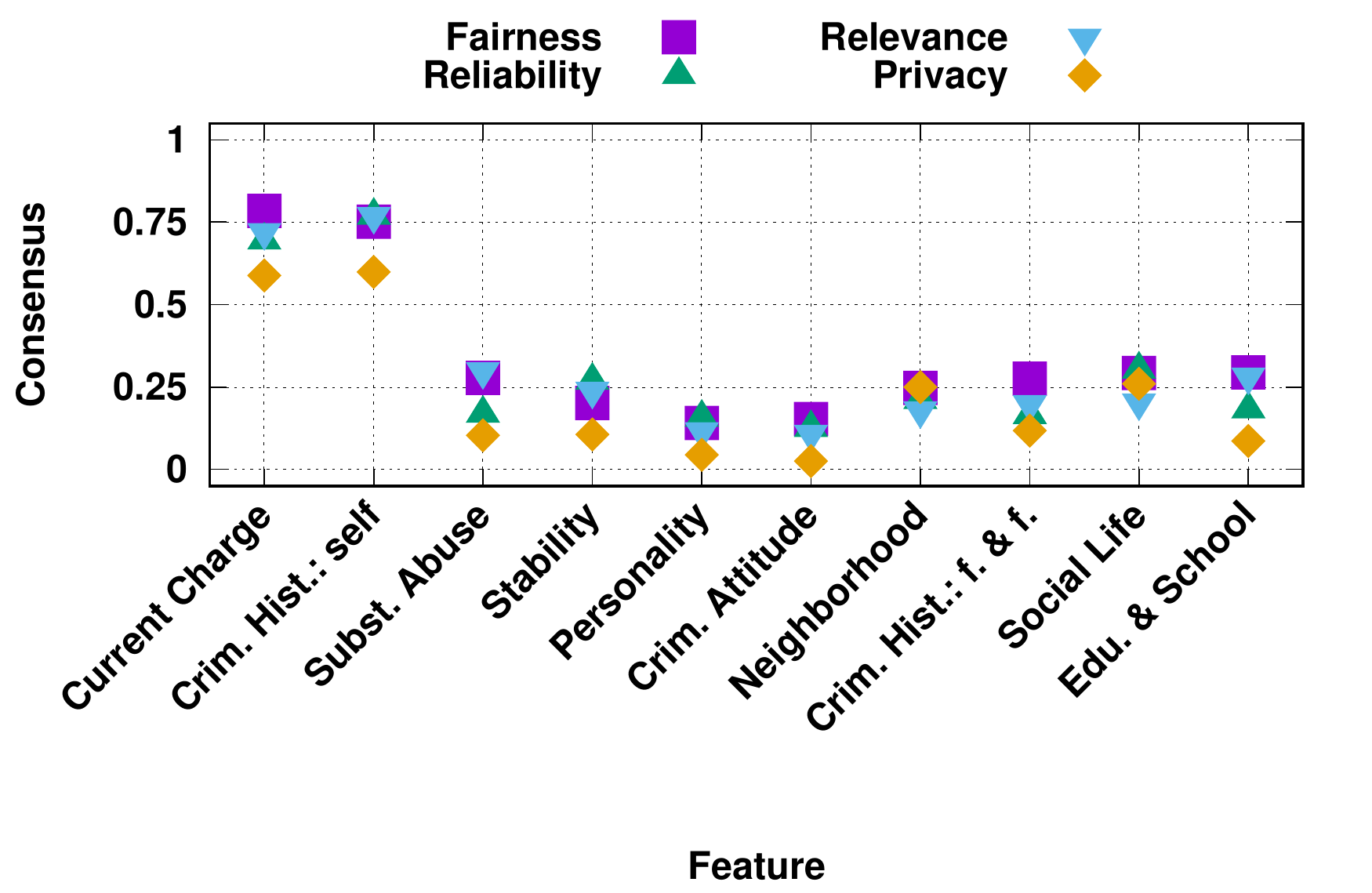}}
\subfloat{\includegraphics[height=0.95\columnwidth,angle=-90]{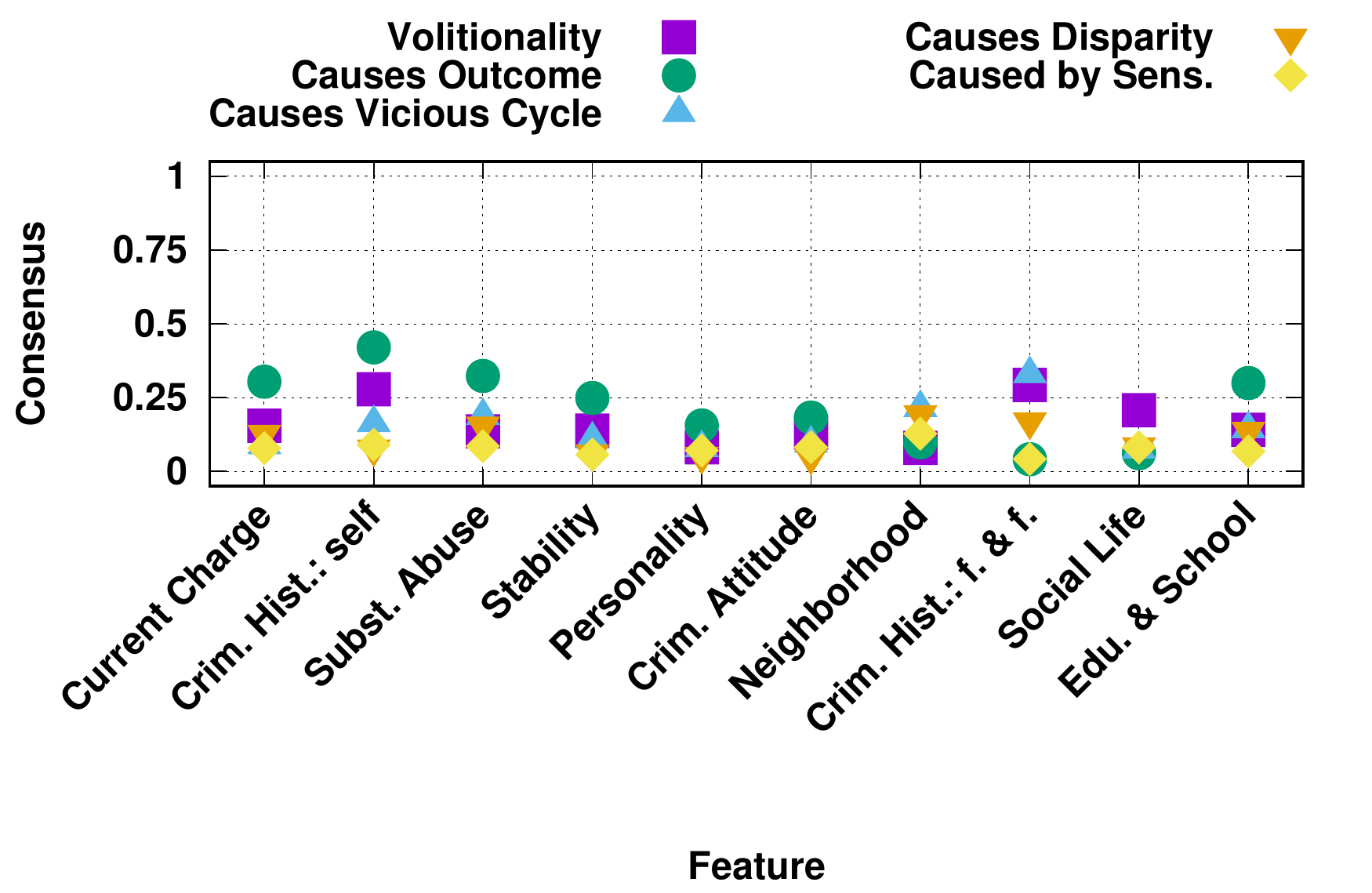}}
\vspace{-2ex}

\caption{\small Consensus in fairness judgments and assessments of latent properties, for the AMT sample. The plots show values of consensus on fairness and the latent properties:
[Left] which exhibit a higher degree of consensus, [Right] which exhibit a lower degree of consensus.}
\label{fig:consensus:properties:AMT}
\vspace{-2ex}

\end{figure*}

\section {Analyzing Fairness Judgments} \label{sec:fairness:ratings}

We first compare respondents' judgments on the fairness of using
different features in our algorithmic decision-making scenario, and
then explore the degree to which they reach consensus in their
judgments on the usage of any given feature.  Throughout the paper, we
conduct our analysis on two datasets: responses of AMT master workers,
which are of higher quality, and responses gathered by SSI, which are
less demographically biased.  Since both datasets exhibit similar
trends, we describe the AMT results in
detail, including only
a brief comparison of the results between the two samples\footnote{For
  the full results on the SSI dataset, please see
  \url{https://fate-computing.mpi-sws.org/procedural_fairness/}}.

\xhdr{Across Different Features} We find that some features are, on average, considered more fair to
use than others.  As shown in Table~\ref{table:fairness:judgments},
AMT respondents rate \textit{Current Charges} and \textit{Criminal
  History} as mostly \textit{fair} to use, with mean ratings close to
6.4.  On the other hand, \textit{Education \& School Behavior},
\textit{Quality of Social Life}, and \textit{Criminal History of
  Family and Friends}, are rated as somewhat \textit{unfair} to use,
with mean ratings close to 2.7.  The remaining features have more
neutral mean ratings, between slightly unfair (rating 3) and slightly
fair (rating 5).

Table~\ref{table:fairness:judgments} also shows that more than half of
the respondents judged five of the ten features as unfair (i.e., gave
a fairness rating between 1 and 3) to be used in this decision-making
scenario. However, as none of these features directly capture
sensitive group membership information such as race or gender of the
defendant, their use for risk prediction is not restricted by
anti-discrimination laws. Thus, our findings suggest that unfairness
concerns about algorithmic decisions are much broader than just
concerns about discrimination.

\xhdr{Across Different Users} Next, we analyze the extent to which
fairness judgments related to the use of any feature vary across the
respondent population. Upon closer examination, we note that
respondents achieve high consensus on only two of the ten features.

In Table \ref{table:fairness:judgments}, we observe that the features
\textit{Current Charges} and \textit{Criminal History} achieve high
consensus, with 95\% and 94\% of respondents, respectively, agreeing
that it is fair to use these features in the algorithmic
decision-making process.  For many of the remaining features, there is
a reasonable consensus amongst respondents, with a two-thirds majority
considering the feature either unfair (by assigning ratings between 1
and 3) or unfair (by assigning ratings between 5 and 7).  However, for
\textit{Personality} and \textit{Criminal Attitudes}, we see very low
consensus, with neither ``fair'' nor ``unfair'' receiving more than a
$>51\%$ slender majority vote.

Table \ref{table:fairness:judgments} also shows consensus in fairness
judgments for different features measured using normalized Shannon
entropy (1-NSE). We observe that features with mean ratings close to
neutral (rating 4) such as \textit{Personality} and \textit{Criminal
  Attitudes} exhibit little consensus, with judgments of respondents
spread across the entire rating spectrum from 1 through 7. Perhaps
surprisingly, respondents also exhibit low consensus on features rated
as least fair to use, such as \textit{Education \& School
  Behavior}. It is possible that societal consensus for or against the
use of these features is still evolving, unlike the broad consensus
against using features like race or gender that has been codified in
anti-discrimination laws.

\xhdr{Impact of Sample Populations} The ranking of features with
respect to mean fairness is similar across AMT and SSI respondents.
However, we note that SSI respondents, in general, reach less
consensus on their fairness ratings than AMT respondents. A possible
explanation is that compared to AMT workers, SSI respondents represent
a more-random and diverse subsample of the general population, and
thus report a wider range of opinions.

\xhdr{Summary} A majority of our survey respondents judged half of the
features used by the COMPAS tool as unfair to use in predicting a
defendant's risk of criminal activity. None of the features directly
capture sensitive group information such as race or gender,
highlighting the need to account for unfairness considerations beyond
those related to discrimination, the sole consideration for most
existing works on algorithmic fairness. However, our findings also
suggest that societal consensus around these other unfairness
considerations might be considerably less evolved than those around
discrimination.

\section{Analyzing Fairness Reasoning} \label{sec:reasoning_behind_fairness}

In this section, we explore the possible causes of the lack of
consensus in respondents' fairness judgments observed in
Section \ref{sec:fairness:ratings}. To this end, we leverage the eight latent properties that we
outlined in Section \ref{sec:framework} as the heuristic basis for how
people judge the fairness of using a feature. Specifically, we will
first examine how people assess the latent properties for different
features and then analyze how the latent property assessments can be
mapped to (i.e., used to predict) fairness judgments. In the process,
we hope to attribute the disagreements in fairness judgments to either
(i) disagreements in the latent property assessments by respondents or
(ii) disagreements in the reasoning respondents use to reach fairness
judgments from latent properties.

\subsection{Latent Property Assessments}\label{subsec:latent_properties}

In Figure \ref{fig:consensus:properties:AMT}, we compare the degree of
consensus in how respondents assessed the eight
properties for different COMPAS input features. In general, we observe
that people tend to disagree in their assessments of all latent
properties for at least one or more features.

However, a closer look reveals important differences.  First,
evaluations of most latent properties related to causality, namely
\textit{causes vicious cycles}, \textit{caused by sensitive features},
\textit{causes disparity in outcomes}, \textit{causal relationship
  with outcomes}, and {\it volitionality}, appear particularly
controversial and exhibit low consensus (< 0.5) for all input features
(shown in the right graph of Figure
\ref{fig:consensus:properties:AMT}.) The consensus around latent
properties related to any feature's potential to cause discrimination,
namely \textit{caused by sensitive features} and \textit{causes
  disparity in outcomes}, is even lower (< 0.2). Our observations have
important implications for recent proposals to avoid algorithmic
discrimination through causal reasoning~\cite{bonchi2017exposing,kilbertus2017avoiding,kusner2017counterfactual,nabi2017fair,whenworlds,qureshi2016causal,zhang2017anti}. These works often assume
that a causal graph, representing causal relationships between
features and outcomes is given as an input to the algorithms. However,
our findings suggest that getting people to agree on a single causal
graph would be a non-trivial challenge.

Second, evaluations of other latent properties not based on causality,
namely \textit{relevance}, \textit{reliability}, and \textit{privacy},
achieve high consensus (> 0.5) on at least some input features (shown
in the left graph of Figure \ref{fig:consensus:properties:AMT}.)
Additionally, we observe that high consensus in these latter latent
property estimates corresponds to high consensus in respondents'
fairness judgments over the corresponding input features.

Thus, we find that (a) our respondents tend to disagree more in their
assessments of causal relationships between input features and
predicted outcomes than in their assessments of non-causality related
latent properties, and (b) consensus, or lack thereof, in certain
latent property assessments appears to be strongly correlated with
consensus in fairness judgments. In the following section, we quantify
the predictive power of the latent property assessments on fairness
judgments.

\xhdr{Impact of Sample Populations} We omit the plot on consensus in
latent property assessments of our SSI respondents due to space
constraints. But, similar to our findings about consensus in fairness
judgments, we find that SSI respondents reach less consensus in their
latent property assessments than AMT respondents.

\subsection{Modeling \& Predicting Fairness Judgments}\label{subsec:fairness_prediction}

We now attempt to model fairness reasoning of our respondents by
attempting to predict their fairness judgments about an input feature
based only on their assessments of the feature's latent
properties. Our insight is as follows: If all our respondents used a
similar reasoning to arrive at fairness judgments from their latent
property estimates, we should be able to learn a single predictor
(mapping function) that would perform well for most, if not all,
respondents. If different respondents used different reasonings, then
no single predictor would perform well for most respondents. Finally,
if we failed to learn predictors that would perform well at the level
of individual users, one could question the validity of our entire
fairness reasoning framework (discussed in
Section~\ref{sec:framework}.)

\begin{table}[t]
\centering
\begin{tabulary}{\linewidth}{C|C|R|R}
	\hline
	\textbf{Sample} &
	\textbf{Neutral Judgments} & \textbf{Accuracy} & \textbf{AUC} \\
	\hline
	{\multirow{2}{*}{AMT}}
	& included & 0.882 & 0.879 \\
	 & excluded & 0.905 & 0.904 \\
	\hline
	{\multirow{2}{*}{SSI}}
	& included & 0.872 & 0.816 \\
	 & excluded & 0.878 & 0.852 \\
	\hline
\end{tabulary}
\caption{\small Accuracy and AUC of binary classifier predicting feature-fairness judgments based on latent property ratings.
Binary classification was performed once with assigning neutral ratings to the fair class in the ground truth, and once excluding neutral fairness judgments from the data, for each sample.}
\label{table:prediction:accuracy}
\vspace{-4ex}
\end{table}

We train a binary logistic regression classifier, as described in
Section \ref{subsec:method_analysis}, and report the results in Table
\ref{table:prediction:accuracy}.  For both datasets, the classifier
achieves very high accuracy (88\% for AMT and 87\% for SSI) when
predicting respondents' fairness judgments based on the underlying
property ratings they assigned.

\begin{table}[ht]
\centering

\begin{tabulary}{\linewidth}{L|CCCCCCC}
	\hline
	\textbf{Rating} & \textbf{1} & \textbf{2} & \textbf{3} & \textbf{4} & \textbf{5} & \textbf{6} & \textbf{7} \\
	\hline
	\textbf{\# Judgments} & 391 & 249 & 195 & 136 & 321 & 280 & 388 \\
	\textbf{\% Misclassified} & 0.06 & 0.16 & 0.36 & 0.33 & 0.09 & 0.04 & 0.01  \\
	\textbf{Avg P Correct} & 0.91 & 0.78 & 0.60 & 0.61 & 0.82 & 0.91 & 0.98 \\
	\textbf{Std P Correct} & 0.19 & 0.26 & 0.30 & 0.31 & 0.22 & 0.16 & 0.08 \\
	\hline
\end{tabulary}
\caption{\small Characterization of misclassifications of our model, by fairness rating value, for the AMT sample.
The first row shows the total number of entries that received a certain rating, while the second row shows the fraction of those instances that were misclassified by our model.
The third and fourth row report on the average value and standard deviation of the probability of being assigned to the correct class respectively.}
\label{table:mification:characterization}
\vspace{-4ex}
\end{table}

In Table \ref{table:mification:characterization}, we examine the few
misclassifications of our model further by studying how they are
distributed over the ground-truth fairness ratings provided by AMT
respondents.  We observe that for features that were rated as very
unfair (1, 2) or very fair (6, 7), our model makes even fewer
mistakes.  On the other hand, for neutral ratings (4), it performs
close to random.  In fact, the average probability that the classifier
predicts the correct class rating for a response and its standard
deviation -- which can be interpreted as the confidence that the model
has in its predictions -- shows that our model is not only highly
accurate
but it is also fairly well calibrated. That is, it is highly confident
(and highly accurate) in its predictions for very unfair or very fair
judgments, but it has low confidence (and low accuracy) in its
predictions for neutral (neither unfair nor fair) judgments.
We hypothesize that these neutral ratings are difficult to predict, as
people may not have clear reasoning underlying their judgments.

\begin{figure}[ht]
\centering
\vspace{-4ex}

\subfloat{\includegraphics[height=0.99\columnwidth,angle=-90]{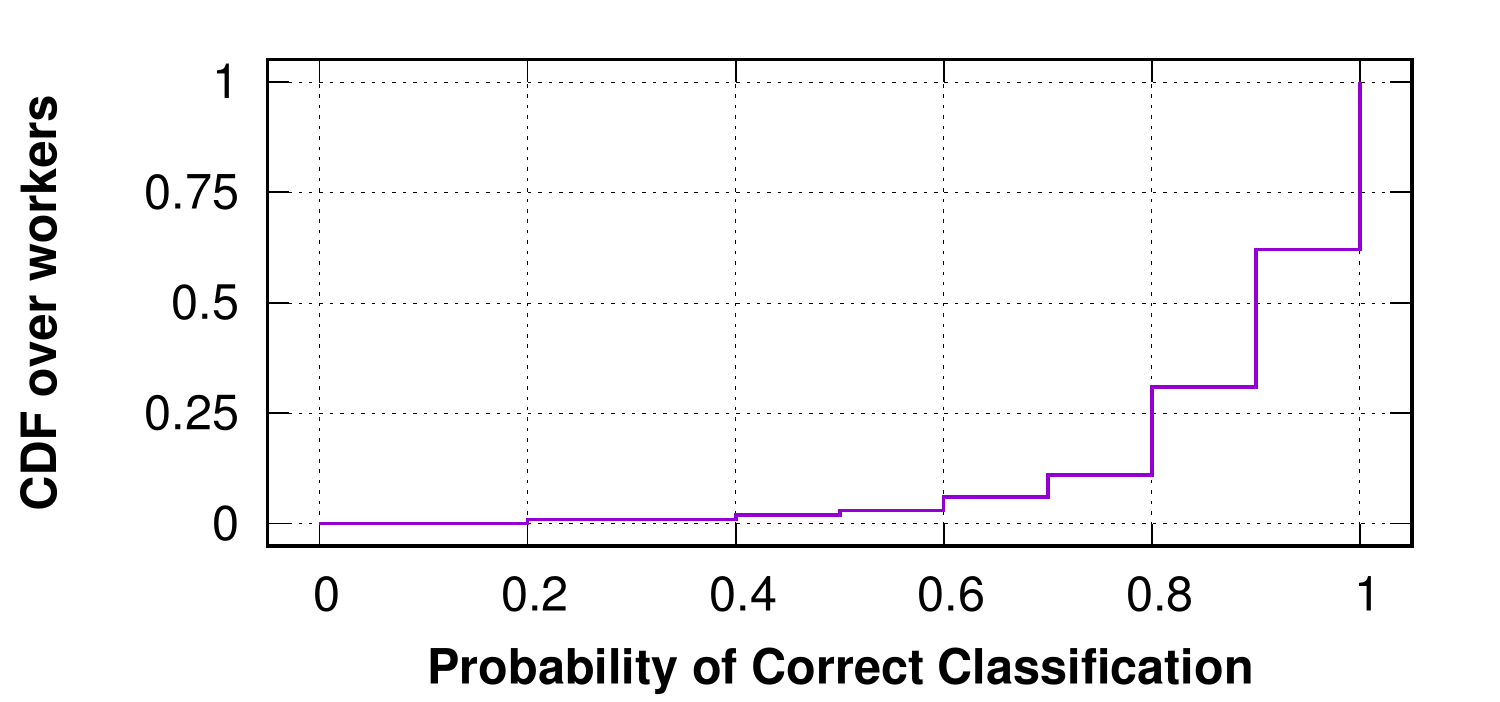}}

\vspace{-2ex}

\caption{\small Characterization of misclassifications of our model, per respondent. The plot shows the CDF of the probability distribution of mistakes over respondents.}

\label{fig:mification:characterization:CDFs}

\end{figure}

Finally, we check if our classifier offers highly accurate predictions
for most respondents. Figure
\ref{fig:mification:characterization:CDFs} shows a cumulative
distribution of inaccurate predictions across the population of
respondents.  We see that our single classifier can make highly
accurate (>= 80\% accuracy) for most (> 85\%) of all our AMT
respondents.

These results suggest that our proposed framework is quite effective
at modeling people's moral judgments about fairness of feature use.
The high accuracy of our predictions for most respondents strongly
suggests (i) that our eight latent properties are largely sufficient
to explain how users arrive at their fairness judgments and (ii) that
most respondents are using similar reasoning at reaching their
fairness judgments from latent properties, even as they disagree on
the assessments of the latent properties in the first place.

\xhdr{Relative Impact of Latent Properties on Fairness Judgments} In
order to estimate the relative importance of our latent properties on
respondents' fairness judgments, we study the odds ratios for the
logistic regression model described above. In this model, we also
include as factors the survey sample source (AMT or SSI) and a mixed-effects term to
account for the multiple judgments provided by each person in our
sample. Table~\ref{tab:or:fairness} summarizes our results.

\begin{table}[ht]
\centering
\begin{tabulary}{\linewidth}{L|C|C|C}
  \hline
 \textbf{Latent Property} & \textbf{O.R.} & \textbf{C.I.} & \textbf{p-value} \\
  \hline
Reliability & 1.27 & [1.2, 1.35] & $<$0.001* \\
Relevance &  2.47 & [2.32, 2.63] & $<$0.001* \\
Privacy & 0.95 & [0.9, 1] & 0.049* \\
Volitionality & 1.18 & [1.13, 1.25] & $<$0.001* \\
Causes Outcome &  1.29 & [1.21, 1.37] & $<$0.001* \\
Vicious Cycle &  0.84 & [0.79, 0.9] & $<$0.001* \\
Causes Disparity & 0.95 & [0.89, 1.02] & 0.159 \\
Caused by Sensitive & 1.03 & [0.96, 1.09] & 0.429 \\
Sample: SSI &1.54 & [1.29, 1.84] & $<$0.001* \\
   \hline
\end{tabulary}
\caption{\small
Binary logistic regression model with binary fairness rating as the outcome variable. In addition to the latent properties, the model also included as inputs the sample source (SSI vs. AMT) and a mixed-effects term to account for multiple measurements from a single survey respondent. O.R. is the log-adjusted regression coefficient (odds ratio), C.I. is the 95\% confidence interval for the O.R., and p-values $<0.05$ are considered significant as dented by a *.}
\label{tab:or:fairness}
\vspace{-4ex}
\end{table}

After controlling for mixed effects, we find that SSI respondents are
more likely to rate a scenario as fair than AMT respondents. As one
might expect, we find that our respondents were more likely to judge
the use of a feature fair, the more they felt that the feature was
\textit{volitional}, \textit{relevant}, \textit{reliable}, and
\textit{caused the outcome}. On the other hand, respondents were less
likely to judge the use of a feature as fair if they felt that the
feature used was \textit{privacy} sensitive, or resulted in a
\textit{vicious cycle}. We find that, while all of the eight
properties helped in prediction, \textit{caused by sensitive group
  membership} and \textit{causes disparity in outcomes} were not
significantly related to judgments in this particular scenario.
Finally, we find that \textit{relevance} had the strongest effect,
with the odds of respondents rating the scenario as fair increasing
2.47 times for every point higher they rated it as relevant (on a
7-point Likert scale).

Note that our analysis here is meant only to illustrate how our
approach could be used to estimate how people implicitly weigh latent
properties of features when making judgments about using features in
algorithmic decisions. However, we do not expect these specific odds
ratios to necessarily hold in other scenarios.

\subsection{Explaining Fairness Disagreements}
Our findings above strongly imply that fairness disagreements likely
arise out of disagreements in how people assess latent properties of
features rather than how they use the latent properties to reason
about fairness of using the features. To confirm this implication, we
used our model from Subsection \ref{subsec:fairness_prediction} to
predict the fairness of using different features.  Our model is
trained on a random subset of AMT responses, and applied on latent
property estimates of the remaining AMT responses, to compute the
resulting fairness judgments.

\begin{figure}[ht]
\centering

\subfloat{\includegraphics[height=0.95\columnwidth,angle=-90]{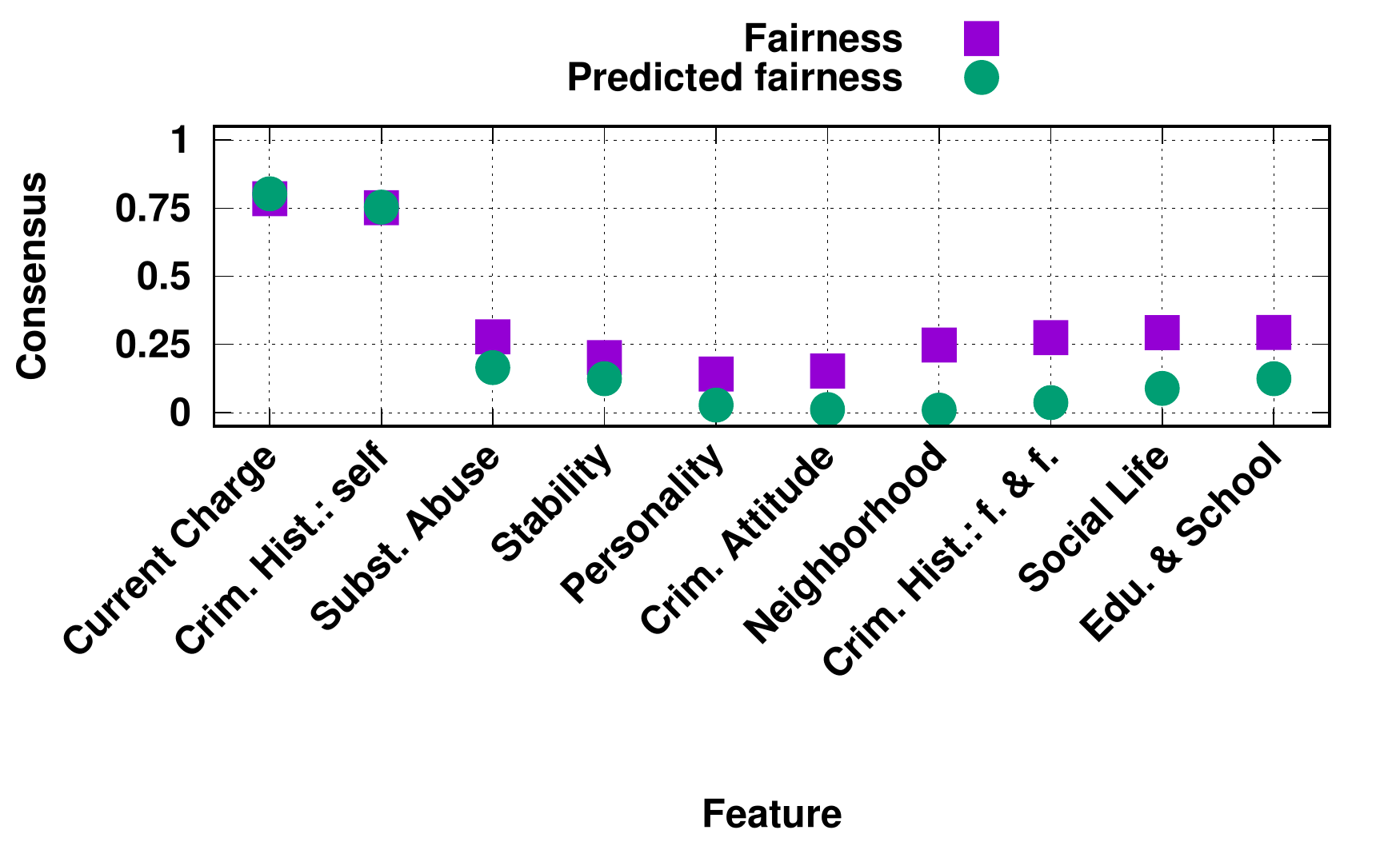}}

\vspace{-2ex}

\caption{\small Consensus in the fairness predictions of our model, compared to the consensus in the ground truth fairness judgments, assigned by people. }
\label{fig:consensus:fairness:predicted}

\end{figure}

Figure \ref{fig:consensus:fairness:predicted} compares the consensus
in the predicted fairness judgments with the consensus in the ground
truth fairness judgments, assigned by respondents.  We observe that
the consensus on predicted fairness ratings tracks the consensus on
ground truth judgments of fairness quite well.  Given that any lack of
consensus in the predicted fairness can be attributed solely to
differences in the assessments of the latent properties, we can
attribute a significant part of the lack of consensus in human
fairness judgments to the differences in people's assessments of
latent properties.

\subsection{Summary}

Overall, our findings validate the framework for reaching fairness
judgments that we proposed in Section~\ref{sec:framework}. We find
significant disagreements in respondents' assessments of many latent
properties, particularly those related to causal relationships between
input features and their causal influence on prediction
outcomes. However, across all respondents we find evidence of strong consensus in the
reasoning
used to reach fairness judgments from
latent properties. Specifically, we showed that we could learn a
single, simple predictor of fairness judgments from latent property assessments that performs with high accuracy across most of our respondents. For the scenario of our survey, the
predictor might be regarded as modeling a common fairness judgment
heuristic used by our respondent population.
We
do not argue that the common heuristic is ``objectively
true''. In fact, we expect the heuristic to depend on cultural
norms of the society. It would be interesting to conduct similar
studies in other societies where people may apply different moral
reasoning to the US population considered here.

\section {Concluding Discussion} \label{sec:discussion}

Most existing works on algorithmic fairness focus on
unfairness due to {\it discrimination}, where people receive
relatively disadvantageous outcomes based on their membership in
certain social groups, e.g., based on race or gender. Further, existing works
take a {\it normative} approach to addressing algorithmic
discrimination, i.e., they prescribe how non-discriminatory decisions
ought to be made.  In contrast, in this work, we take a {\it
  descriptive} approach to algorithmic fairness, i.e., we ask people
what they perceive as unfair in decision making, and analyze the reasons
behind their unfairness perceptions.

Our study focused on how people perceive the unfairness of using
different features describing a defendant to algorithmically predict the
defendant's risk of engaging in criminal activity in the near
future. Our survey study reveals several interesting
findings:
(i) People's concerns about the unfairness of using a feature
extend far beyond discrimination, including consideration of latent properties  such as the relevance of the feature to the decision
making scenario and the reliability with which the feature can be
assessed.
(ii) Unfortunately, there are considerable disagreements on
which features different people perceive as unfair to use.
(iii) The lack of consensus can be attributed to disagreements in how people
assess the latent properties of the features, particularly those
related to causal relationships between input features and their
causal influence on outcomes.
(iv) Encouragingly, different people appear to
share a common heuristic (i.e., a similar reasoning) when reaching
their fairness judgment from their assessments of latent properties.

Our observations yield several implications for future studies on
algorithmic fairness:
(i) While there are important reasons based on historic prejudice to mitigate discrimination, there is strong evidence here to
consider additional unfairness concerns.
(ii) The lack of consensus on causal relationships between input features and outcomes
raises challenges for approaches to fairness based on causal reasoning which require a known causal structure (one recent approach can incorporate multiple structures \cite{whenworlds}).
(iii) On the other hand, our findings
highlight the desirability of trying to gather more objective causal data.
(iv) Such objective data on latent properties, if possible, could then
be used as inputs to a common fairness heuristic (moral reasoning),
which our evidence indicates is shared by most people, to arrive at
consensus fairness judgments. It would be interesting to explore the
extent to which this heuristic varies across different cultures and
decision making contexts.

\bibliographystyle{ACM-Reference-Format}

\end{document}